\documentclass[
 amsmath,amssymb,
 preprint,
 superscriptaddress
]{article}

\usepackage{amsmath}
\usepackage{amssymb}
\usepackage[english]{babel}
\usepackage{graphicx}
\usepackage{bm}
\usepackage{epsfig}
\usepackage[svgnames]{xcolor}
\usepackage{pstricks,graphics,pst-grad,times}
\usepackage{listings}
\usepackage{microtype}
\usepackage{pifont}
\usepackage{authblk}
\usepackage{blindtext}
\usepackage{hyperref}
\usepackage{threeparttable} 
\usepackage{booktabs}       
\title{Benchmarking Synthetic Tabular Data:\\A Multi-Dimensional Evaluation Framework}

\author{Andrey Sidorenko}
\author{Michael Platzer}
\author{Mario Scriminaci}
\author{Paul Tiwald}
\affil{\texttt{\{andrey.sidorenko, michael.platzer, mario.scriminaci, paul.tiwald\}@mostly.ai}\vspace{2ex}\\
MOSTLY AI}

\date{}

\begin{document}
\maketitle
\begin{abstract}
Evaluating the quality of synthetic data remains a key challenge for ensuring privacy and utility in data-driven research. In this work, we present an evaluation framework that quantifies how well synthetic data replicates original distributional properties while ensuring privacy. The proposed approach employs a holdout-based benchmarking strategy that facilitates quantitative assessment through low- and high-dimensional distribution comparisons, embedding-based similarity measures, and nearest-neighbor distance metrics. The framework supports various data types and structures, including sequential and contextual information, and enables interpretable quality diagnostics through a set of standardized metrics. These contributions aim to support reproducibility and methodological consistency in benchmarking of synthetic data generation techniques. The code of the framework is available at \hyperlink{https://github.com/mostly-ai/mostlyai-qa}{https://github.com/mostly-ai/mostlyai-qa}.

\end{abstract}


\section{Introduction}

Generative Artificial Intelligence (AI) is rapidly transforming data-centric research fields, transcending its initial prominence in unstructured data domains, such as natural language processing and image synthesis, to structured and semi-structured data contexts prevalent within organizational data assets. Synthetic data generation specifically addresses critical challenges, including privacy-preserving data sharing, representation enhancement of underrepresented subpopulations, simulation of rare but consequential scenarios, and imputation of missing data \cite{assefa2020generating, jordon2022synthetic, van2024synthetic}. However, the practical utility and acceptance of generative synthetic data critically depend on a rigorous evaluation of its fidelity (accuracy of representation) and novelty (degree of originality).

Despite the existence of numerous evaluation frameworks for synthetic data \cite{howe2017synthetic, lu2019empirical, platzer2021holdout, chundawat2022universal, chundawat2022tabsyndex, alaa2022faithful, espinosa2023quality, sdnist, hudovernik2024benchmarking, cannon2025assessinggenerativemodelsstructured}, comprehensive and accessible tools addressing both fidelity and novelty simultaneously remain scarce. See Table~\ref{tab:tool_comparison} for a high-level tool comparison. In particular, existing tools often emphasize one evaluation dimension at the expense of the other, yielding either high fidelity through replication or high novelty through randomness, but rarely balancing the two dimensions effectively. For instance, merely copying original samples yields high accuracy without being novel, while generating entirely random samples scores high on novelty without being accurate. The true challenge of privacy-safe synthetic data lies in the generation of data that is both accurate \emph{and} novel. Thus, any quality assurance for synthetic data must measure \emph{both} of these dimensions.
\renewcommand{\arraystretch}{1.2}
\begin{table}[h]
    \centering
    \begin{threeparttable}
    \resizebox{\textwidth}{!}{%
    \begin{tabular}{|l|l|c|c|c|c|c|}
        \hline
        \textbf{Python package} & \textbf{License} & \textbf{HTML} & \textbf{Plots} & \textbf{Metrics} & \textbf{Novelty} & \textbf{Data} \\ \hline
        \texttt{mostlyai-qa} (2024)\tnote{a} & Apache  & \checkmark & \checkmark & \checkmark & \checkmark & flexible  \\ \hline
        \texttt{ydata-profiling} (2023)\tnote{b} & MIT  & \checkmark & \checkmark & -- & -- & flexible   \\ \hline
        \texttt{sdmetrics} (2023)\tnote{c} & MIT   & --  & \checkmark & \checkmark & \checkmark & flexible  \\ \hline
        \texttt{synthcity} (2023)\tnote{d} & Apache    & --  & \checkmark & \checkmark & \checkmark & flexible  \\ \hline
        \texttt{sdnist} (2023)\tnote{e} & Permissive  & \checkmark  & \checkmark & \checkmark & $\sim$ & fixed  \\ \hline
    \end{tabular}
    } 
    \begin{tablenotes}
        \footnotesize
        \item[a] \url{https://github.com/mostly-ai/mostlyai-qa}
        \item[b] \url{https://github.com/ydataai/ydata-profiling}
        \item[c] \url{https://github.com/sdv-dev/SDMetrics}
        \item[d] \url{https://github.com/vanderschaarlab/synthcity}
        \item[e] \url{https://github.com/usnistgov/SDNist}
    \end{tablenotes}
    \caption{Comparison across open-source Python libraries for assessing synthetic data.}
    \label{tab:tool_comparison}
    \end{threeparttable}
\end{table}
To fill this methodological void, we introduce \texttt{mostlyai-qa}, an open-source Python framework explicitly designed to comprehensively evaluate the quality of synthetic data. The framework uniquely integrates accuracy, similarity, and novelty metrics within a unified evaluation framework. It effectively handles diverse data types, including numerical, categorical, datetime, and textual, as well as data with missing values and variable row counts per sample, accommodating multi-sequence, multivariate time-series data\footnote{Multi-sequence time-series data is the predominant structure for behavioral data, where multiple events for multiple individuals are recorded.}. The quality of synthetic data is also evaluated by taking into account any contextual data.

The primary contributions of this paper include:
\begin{itemize}
\item A novel evaluation framework that simultaneously assesses fidelity and novelty of synthetic datasets.
\item Support for comprehensive, automated assessment and visualization of mixed-type data quality.
\item Open-source availability under the Apache License v2, promoting broad adoption and collaborative enhancement within the research community.
\end{itemize}

\section{A framework for evaluation of synthetic data}

The evaluation of synthetic data requires careful consideration of two primary dimensions: fidelity and novelty. Fidelity describes the degree to which synthetic samples represent the statistical properties of original data, while novelty ensures that generated samples are distinct enough to preserve privacy and avoid direct replication. The framework combines these concepts by employing an empirical holdout-based assessment for synthetic mixed-type data, introduced in \cite{platzer2021holdout}. In that framework, the quality of synthetic data is benchmarked against holdout data samples that were not used in privacy-preserving training, expecting models to produce novel samples that reflect the underlying data distribution without direct replication. Accordingly, synthetic samples should be as close to training samples as holdout samples but not closer. This approach, akin to the use of holdout samples for supervised learning, enables the evaluation of a generative model's ability to generalize underlying patterns rather than merely memorizing specific training samples.
\begin{figure}[!ht]
\centering
\includegraphics[width=0.8\textwidth]{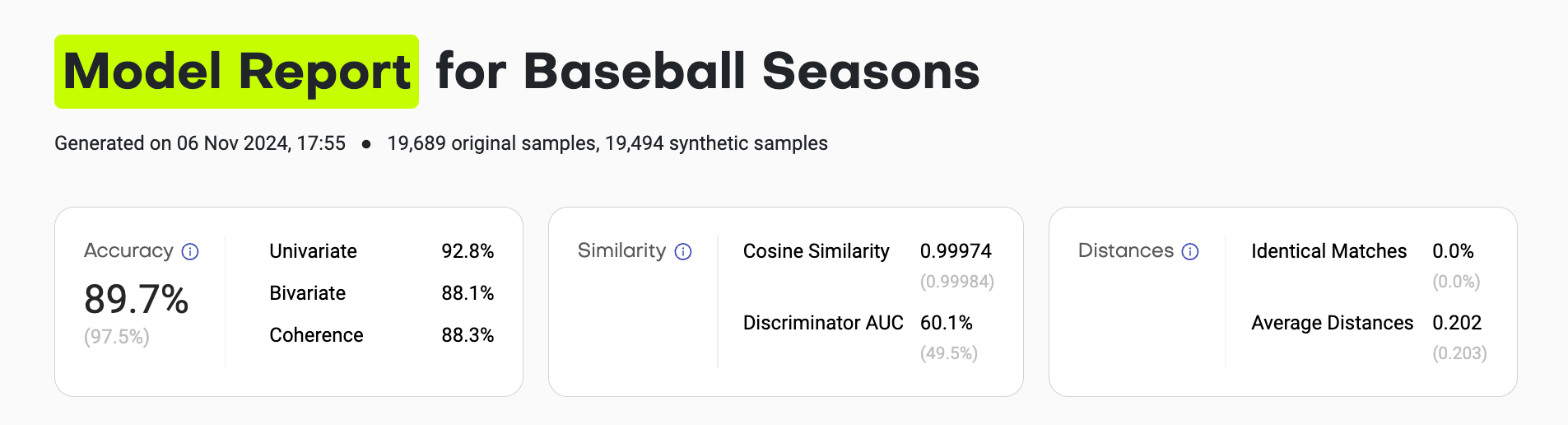}
\caption{An example of metrics summary generated by the framework.}
\label{fig:metrics}
\end{figure}
The framework is built upon that framework and structured around three interrelated categories of metrics - Accuracy, Similarity, and Distances - each comprising specific submetrics that collectively address the dual objectives of fidelity and novelty. Accuracy quantifies lower-dimensional, and similarity higher-dimensional fidelity, whereas the set of distance metrics helps to gauge the novelty of samples (see Fig.~\ref{fig:metrics}).

\subsection{Accuracy}\label{sec:accuracy}

Accuracy metrics assess how closely synthetic data replicate the low-order marginal, joint distributions, and consistency along the time dimension (sequential data coherence) of the original dataset, with a score of 100\% representing an exact match. The overall accuracy score is computed as 100\% minus the top-k total variation distance (with k=10) aggregated across three components:
\begin{itemize}
    \item \textbf{Univariate accuracy}: Measures fidelity of discretized univariate distributions across all attributes.
    \item \textbf{Bivariate accuracy}: Captures alignment between pairs of attributes via discretized bivariate frequency tables.
    \item \textbf{Coherence}: Evaluates attribute consistency across sequential records, applicable only to sequential data.
\end{itemize}

\begin{figure}[!ht]
\centering
\includegraphics[width=0.8\textwidth]{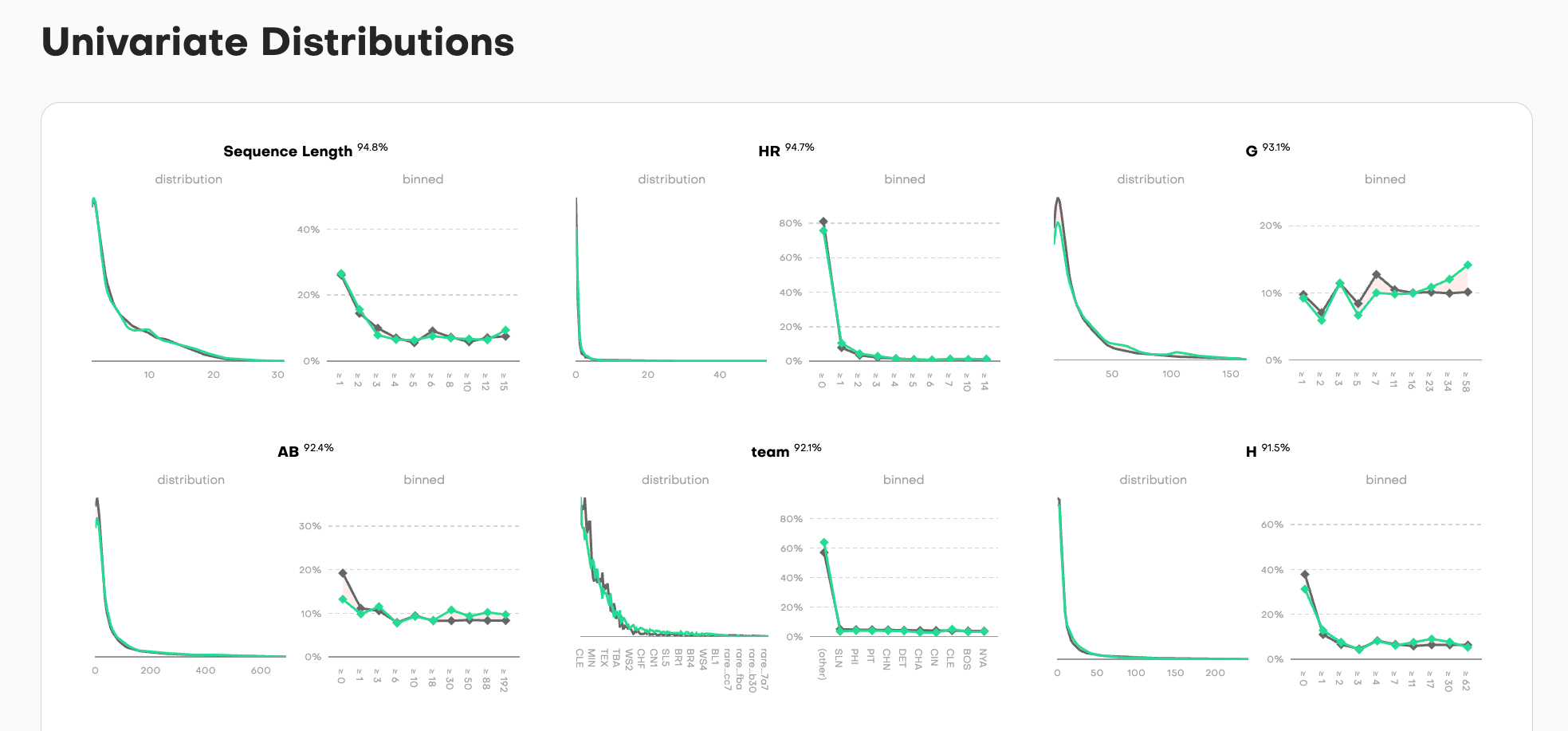}
\caption{An example of univariate distributions and their accuracies generated by the framework.}
\label{fig:univariates}
\end{figure}

\begin{figure}[!ht]
\centering
\includegraphics[width=0.8\textwidth]{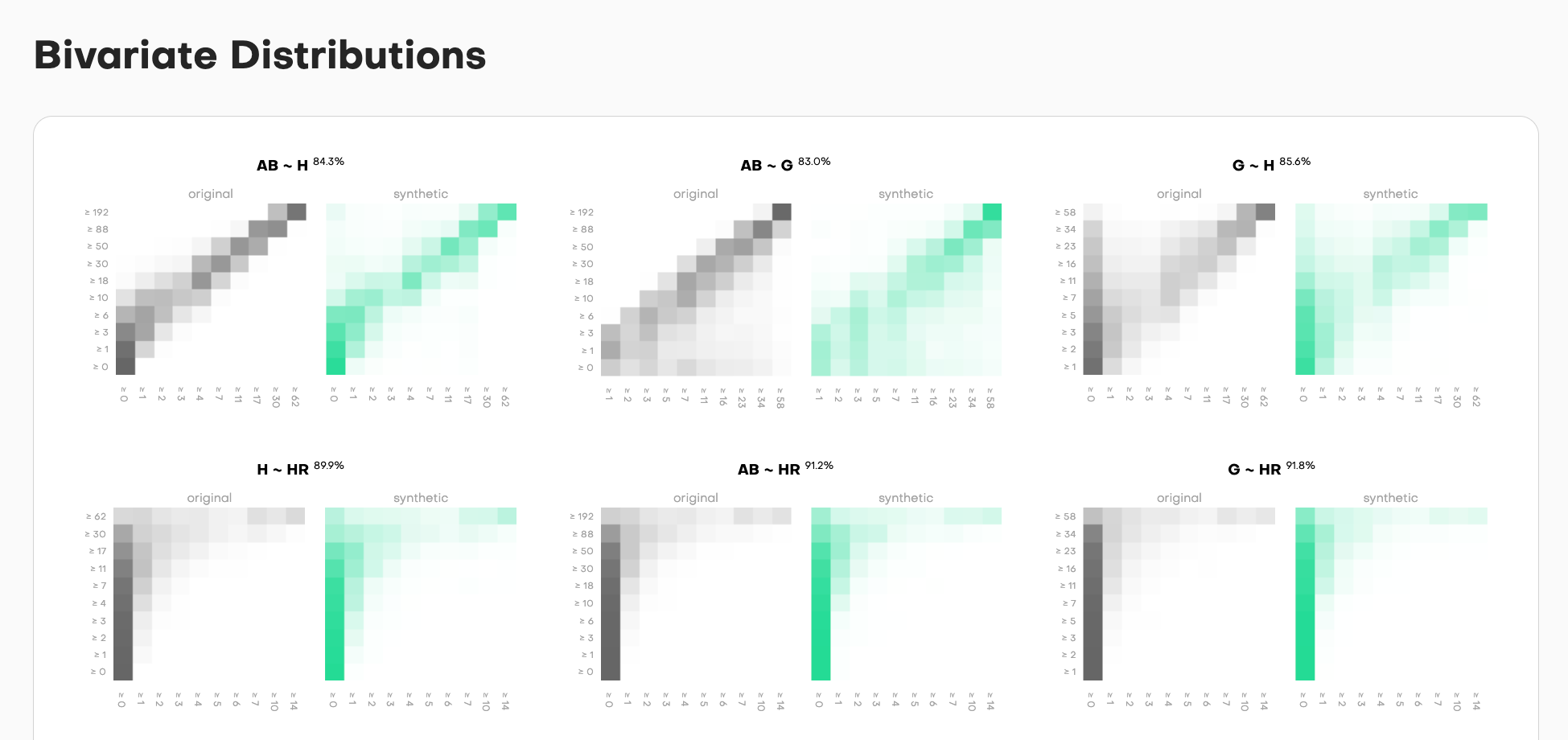}
\caption{Bivariate distributions and their accuracies generated by the framework.}
\label{fig:bivariates}
\end{figure}

To evaluate low-order marginals, univariate distributions (Fig.~\ref{fig:univariates}) and pairwise correlations between columns (bivariate distributions; Fig.~\ref{fig:bivariates}) are compared. For datasets containing mixed data types, numerical and date-time columns are transformed by discretizing their values into deciles based on the original training data, creating ten equally populated groups per column. For categorical columns, only the ten most frequent categories are retained, while the less common ones are excluded. This method enables a consistent comparison across different data types, emphasizing the most informative features of the data.
For each feature, we derive two vectors of length 10, one from the original training data and one from the synthetic data. In the case of numerical and date-time columns, these vectors capture the frequency of values within the decile-based groups defined by the original data. For categorical columns, the vectors represent the re-normalized frequency distribution of the top ten most frequent categories. These feature-specific vectors are denoted as $\mathbf{X}_{\text{trn}}^{(m)}$ and $\mathbf{X}_{\text{syn}}^{(m)}$, corresponding to the training and synthetic data, respectively. $m$ is the feature index, running from 1 to $d$.

The \textbf{univariate accuracy} of column $m$ is then defined as
\begin{equation}
    acc_{\text{univariate}}^{(m)} = \frac{1}{2} \left(1 - \|\mathbf{X}_{\text{trn}}^{(m)} - \mathbf{X}_{\text{syn}}^{(m)}\|_1 \right)
\end{equation}
and the overall univariate accuracy, as reported in the results section, is defined by
\begin{equation}
    acc_{\text{univariate}} = \frac{1}{D} \sum_m^D acc_{\text{univariate}^{(m)}} \; ,
\end{equation}
where $D$ is the number of columns.

For bivariate metrics, the relationships between pairs of columns are assessed using normalized contingency tables. These tables represent the joint distribution of two features, $m$ and $n$, enabling the evaluation of pairwise dependencies.

The contingency table between columns $m$ and $n$ is denoted as $\mathbf{C}_{\text{trn}}^{(m,n)}$ for the training data and $\mathbf{C}_{\text{syn}}^{(m,n)}$ for the synthetic data. Each table has a maximum dimension of 10×10, corresponding to the (discretized) values or the top ten categories of the two features. For columns with fewer than ten categories (categorical columns with cardinality \textless 10), the dimensions of the table are reduced accordingly.

Each cell in the table represents the normalized frequency with which a specific combination of categories or discretized values from columns $m$ and $n$ appears in the data. This normalization ensures meaningful comparisons across different features and datasets, independent of their original scale or size.

The \textbf{bivariate accuracy} of the column pair $m,n$ is then defined as
\begin{align}
    acc^{(m,n)}_{\text{bivariate}} &= \frac{1}{2}\left(1 - \|\mathbf{C}_{\text{trn}}^{(m,n)} 
    - \mathbf{C}_{\text{syn}}^{(m,n)}\|_{1,\text{entrywise}}\right) \nonumber \\
    &= \frac{1}{2}\left(1 - \sum_i \sum_j 
    \left|\mathbf{C}_{\text{trn}}^{(m,n)} - \mathbf{C}_{\text{syn}}^{(m,n)}\right|_{i,j}\right)
\end{align}
and the overall bivariate accuracy, as reported in the results section, is given by
\begin{equation}
    acc_\text{bivariate} \frac{2}{D(D-1)} \sum_{1 \leq m < n \leq D} acc_\text{bivariate}^{(m,n)} \; ,
\end{equation}  
the average of the strictly upper triangle of $acc_\text{bivariate}^{(m,n)}$.

Note that due to sampling noise, both $acc_{\text{univariate}}$ and $acc_{\text{bivariate}}$ cannot reach 1 in practice. The software package reports the theoretical maximum alongside both metrics.

There is no difference in calculating the univariate and bivariate accuracies between flat and sequential data. In both cases, the vectors $\mathbf{X}^{(m)}$ and contingency tables $\mathbf{C}^{(m,n)}$ are based on all entries in the columns, irrespective of which data subject they belong to.
\begin{figure}[!ht]
\centering
\includegraphics[width=0.8\textwidth]{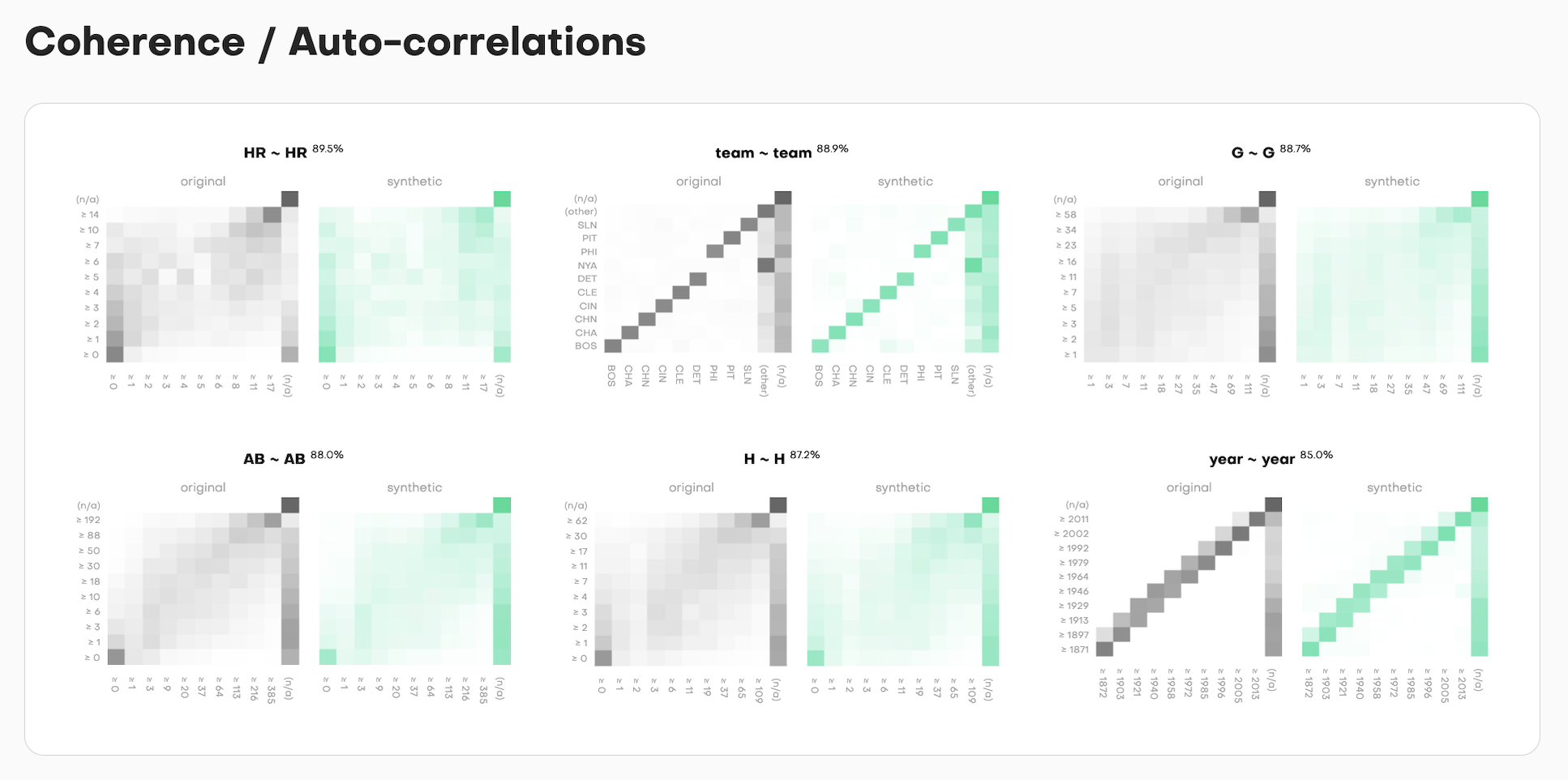}
\caption{An example of coherence distributions and their accuracies generated by the framework.}
\label{fig:coherences}
\end{figure}
For sequential data, the framework evaluates the consistency (coherence) of relationships between successive time steps or sequence elements (Fig.~\ref{fig:coherences}). This allows the assessment of whether the original sample autocorrelations within sequences are faithfully reproduced in the synthetic data. The process is as follows:
\begin{itemize}
    \item For each data subject, we randomly sample two successive sequence elements (time steps) from their sequential data.
    \item These pairs of successive time steps are transformed into a wide-format dataset. To illustrate, consider a sequential dataset of $N$ subjects and original columns $A, B, C$, represented as $K>N$ rows. After processing, the resulting dataset has six columns: $A, A', B, B', C, C'$. The unprimed columns correspond to the first sampled sequence element, the primed columns correspond to the successive sequence element. The number of rows in this wide-format dataset is equal to $N$, irrespective of the sequence lengths in the original dataset.
\end{itemize}
Using this wide-format dataset, we construct contingency tables $\mathbf{C}^{(m,m')}$ for each pair of corresponding unprimed and primed columns $(m,m')$. These tables are normalized and used to calculate the \textbf{coherence metric} for column $m$ as:
\begin{equation}
    acc^{(m,m')}_{\text{coherence}} = \frac{1}{2}\left(1 - \|\mathbf{C}_{\text{trn}}^{(m,m')} - \mathbf{C}_{\text{syn}}^{(m,m')}\|_{1,\text{entrywise}}\right)
\end{equation}
and the overall coherence metric, as reported in the results section
\begin{equation}
    acc_\text{coherence} = \frac{1}{D} \sum_{m}^{D} acc^{(m,m')}_\text{coherence} \; .
\end{equation}

We summarize the \textbf{overall accuracy} of a data set as
\begin{equation}
    \frac{1}{2} \left( acc_\text{univariate} + acc_\text{bivariate} \right)
\end{equation} and
\begin{equation}
    \frac{1}{3} \left( acc_\text{univariate} + acc_\text{bivariate} + acc_\text{coherence} \right)
\end{equation}
for flat and sequential data, respectively.

This approach offers consistency across attribute types. Additionally, the overall accuracy metric is decomposable into 1-way and 2-way frequency tables, which are visualized as density and heat-map plots, respectively, making it easily interpretable also for non-statisticians. The greater the discrepancies between the plotted distributions, the lower the accuracy score. To achieve a high overall accuracy, each contributing distribution must align closely with the original. However, due to sampling noise with finite samples, some discrepancies are inevitable. By calculating the expected accuracy for a theoretical holdout dataset based on the original distributions and sample size, we provide a reference benchmark. Rather than aiming for 100\% accuracy, the goal is for synthetic samples to match this benchmark closely, indicating similarity to the training samples akin to holdout samples.

When contextual data is present, the framework will report the accuracy of bivariate distributions between contextual and target attributes, enabling assessment of whether these relationships are well-preserved in the synthetic data.

\subsection{Centroid Similarity}\label{sec:similarity}

Complementing accuracy, we report another set of metrics that assess the similarity of distributions. Rather than analyzing the easy-to-interpret lower-dimensional marginals, the focus shifts to the high-dimensional full joint distributions. Direct analysis of high-dimensional distributions is not feasible due to the curse of dimensionality, so we use an alternative approach. Every tabular sample is first converted into a string of values (e.g., \texttt{value\_col1;value\_col2;...;value\_colD}), that is then mapped into an informative embedding space using a pre-trained language model. For sequential data, the string is constructed by concatenating the values of all columns across time steps. For instance, values from time step two are appended to the string containing values from time step one, and so on. For long sequences, the resulting input string is truncated to fit within the language model’s context window.

While the choice of language model is flexible, we specifically opted for \texttt{all\-MiniLM\-L6\-v2}\footnote{\href{https://huggingface.co/sentence-transformers/all-MiniLM-L6-v2/}{https://huggingface.co/sentence-transformers/all-MiniLM-L6-v2/}} as it is a lightweight, compute-efficient universal model. It transforms each string of values into a $384$-dimensional embedding space. Then centroids for each group of embeddings are calculated as
\begin{equation}
\mathbf{c}_{\mathrm{syn}} = \frac{1}{n_{\mathrm{syn}}} \sum_{i=1}^{n_{\mathrm{syn}}} \mathbf{x}_{\mathrm{syn}, i}, 
\quad
\mathbf{c}_{\mathrm{trn}} = \frac{1}{n_{\mathrm{trn}}} \sum_{i=1}^{n_{\mathrm{trn}}} \mathbf{x}_{\mathrm{trn}, i}, 
\quad
\mathbf{c}_{\mathrm{hol}} = \frac{1}{n_{\mathrm{hol}}} \sum_{i=1}^{n_{\mathrm{hol}}} \mathbf{x}_{\mathrm{hol}, i},
\end{equation}

where \( \mathbf{X}_{\mathrm{syn}} \in \mathbb{R}^{n_{\mathrm{syn}} \times d} \) is the matrix with rows representing embeddings of synthetic data, 
\( \mathbf{X}_{\mathrm{trn}} \in \mathbb{R}^{n_{\mathrm{trn}} \times d} \) the matrix 
for embeddings of training data, and \( \mathbf{X}_{\mathrm{hol}} \in \mathbb{R}^{n_{\mathrm{hol}} \times d} \) the matrix for holdout embeddings if provided.

We then compare the centroids of the synthetic and training samples using cosine \textbf{similarity}
\begin{equation}
\mathrm{cosine\_similarity}(\mathbf{u}, \mathbf{v}) \;=\;
\frac{\mathbf{u} \cdot \mathbf{v}}{\|\mathbf{u}\| \,\|\mathbf{v}\|},
\end{equation}
aiming for a high similarity score (with an upper bound of 1). However, to account also here for sampling variance, we use the cosine similarity between the training and holdout centroids as a reference, ensuring that synthetic samples are close to the training distribution without exceeding the similarity expected due to natural sampling noise.
\begin{figure}[!h]
\centering
\includegraphics[width=0.8\textwidth]{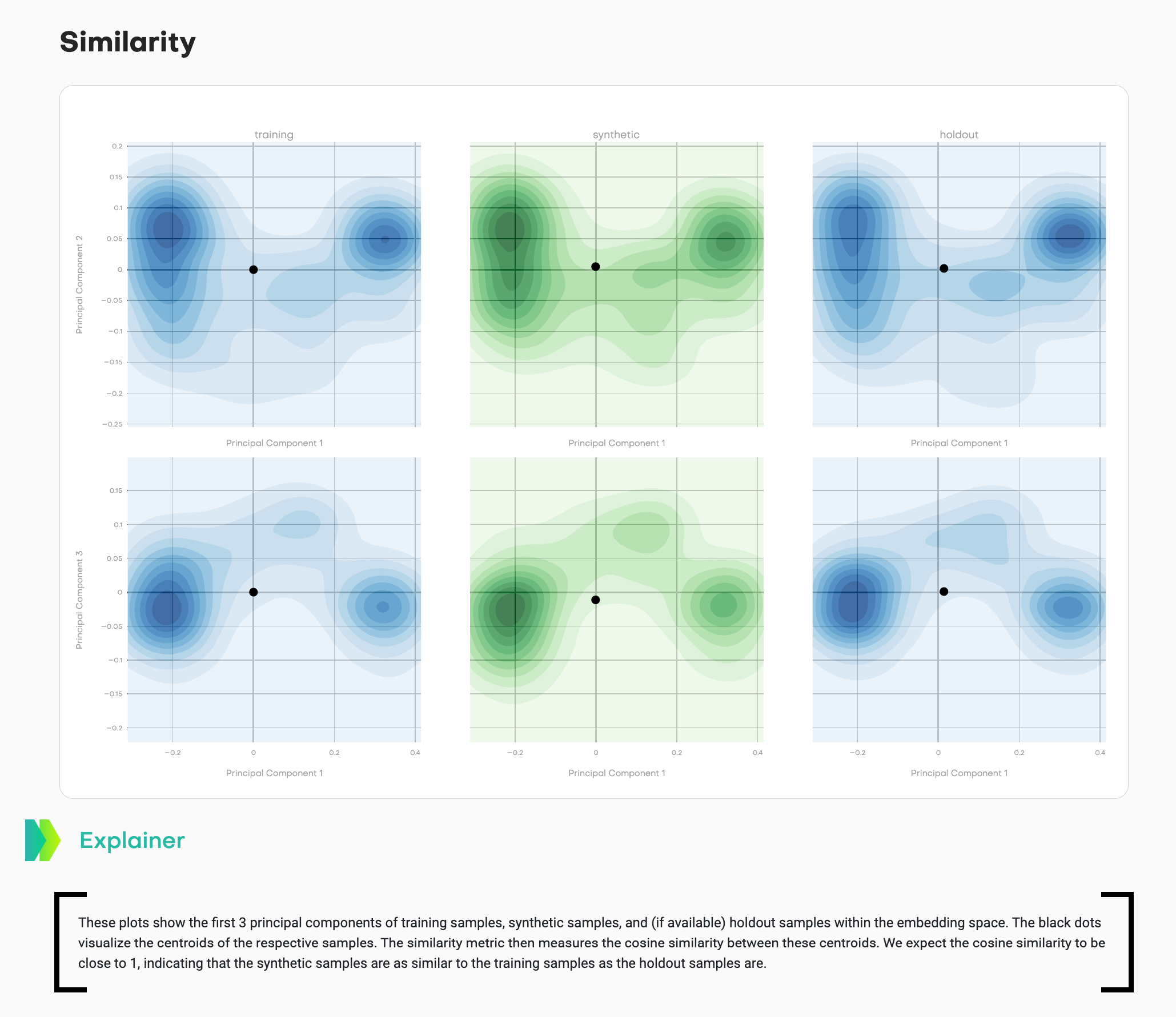}
\caption{Similarity within PCA-projected embedding space generated by the framework.}
\label{fig:similarity}
\end{figure}
To enhance interpretability, we also provide a visualization of the embedded samples and their centroids, projected into a lower-dimensional space using Principal Component Analysis (PCA) (Fig.~\ref{fig:similarity}).

In addition to cosine similarity, we leverage the embedding space to train a discriminative model that indicates whether synthetic samples are truly indistinguishable from training samples. If certain properties of the synthetic samples (e.g., implausible attribute combinations) reveal them as synthetic rather than real, the area-under-the-curve (AUC) metric quantifies this distinguishability.

\subsection{Distances}\label{sec:distances}

Synthetic samples should resemble \emph{novel} samples from the original distribution rather than simply replicating seen samples. Consequently, they are expected to be just as close to training samples as to holdout samples.

\begin{figure}[!ht]
\centering
\includegraphics[width=0.8\textwidth]{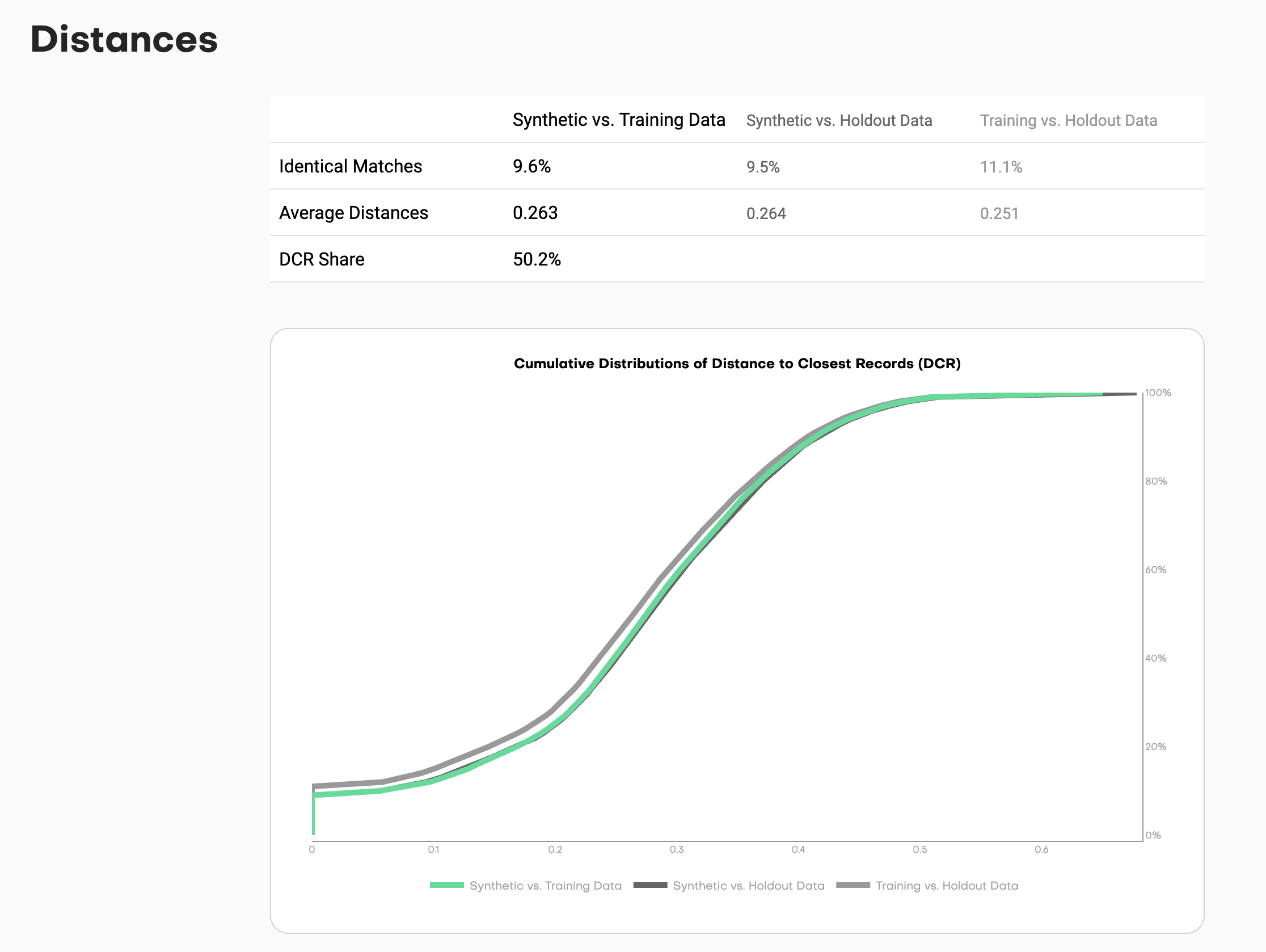}
\caption{Cumulative distributions of distances to closest records (DCRs) for assessing novelty.}
\label{fig:distances}
\end{figure}

Thus, we assess the novelty of synthetic data by examining distances between samples within the high-dimensional embedding space introduced in Section~\ref{sec:similarity}. For each synthetic sample, we calculate the distance to its closest record (DCR) among the training samples. This nearest-neighbor distance is expected to vary depending on whether the sample is a synthetic inlier or outlier. Therefore, absolute distances alone cannot reliably indicate novelty; instead, we need to contextualize these values by comparing them to the same distances calculated for an equally sized holdout dataset. This comparison is performed for both the average DCR, which we report as a metric, and the overall cumulative DCR distribution, which is visualized (Fig.~\ref{fig:distances}). For reference, the average distances between the synthetic records and their nearest neighbors from the holdout dataset are also displayed.

With the sample embeddings denoted as $\text{emb}_i$ and $i$ ranging from $1$ to $N$, the nearest neighbor distances are calculated using the L2 norm between embedded representations of synthetic, training, and holdout records. For an embedded synthetic record $\text{emb}_i^\text{(syn)}$, the distance to its nearest neighbor in the training and holdout datasets is computed as:
\begin{equation}
    d_{\text{trn}}^{(i)} = \min_{j \in N_{\text{trn}}} \|\text{emb}_i^{(\text{syn})} - \text{emb}_j^{(\text{trn})}\|_2, \quad
    d_{\text{hold}}^{(i)} = \min_{j \in N_{\text{hold}}} \|\text{emb}_i^{(\text{syn})} - \text{emb}_j^{(\text{hold})}\|_2.
\end{equation}
With the indicator function
\begin{equation}
\mathbb{I}_{\text{trn}}^{(i)} =
\begin{cases} 
1 & \text{if } d_{\text{trn}}^{(i)} < d_{\text{hold}}^{(i)}, \\
0 & \text{if } d_{\text{trn}}^{(i)} > d_{\text{hold}}^{(i)}, \\
0.5 & \text{if } d_{\text{trn}}^{(i)} = d_{\text{hold}}^{(i)},
\end{cases}
\end{equation}
which indicates whether the nearest neighbor of $\text{emb}_i^\text{(syn)}$ is in the training set, we define the \textbf{DCR share} as
\begin{equation}
\text{DCR share} = \frac{1}{N_{\text{syn}}} \sum_{i=1}^{N_{\text{syn}}} \mathbb{I}_{\text{trn}}^{(i)}.
\end{equation}

It is equally important to compare against the corresponding holdout metrics when evaluating identical matches—instances where synthetic records are exactly the same as the original across all attributes. Crucially, the presence of identical matches does not automatically imply a lack of novelty. If the original data includes duplicates, we should expect (and even require) a similar level of duplication in the synthetic data. Simply removing individual records in an effort to enforce novelty is not only insufficient but could also increase the risk of exposing original data\cite{hann2024}.

\section{Empirical Demonstration}

By splitting the original data into training and holdout samples and, subsequently, generating multiple synthetic datasets based on the training data, we can effectively compare quality across various generation methods. The chart below visualizes key metrics relative to their holdout-based reference metrics for the UCI Adult Census dataset \cite{adult}, as synthesized and published in \cite{platzer2021holdout}. The closer a synthesizer approaches the \emph{north star} reference point at \texttt{(1, 1)} - the holdout data set - the better its privacy-utility trade-off. As illustrated, this trade-off applies to AI-based data synthesizers just as it does to traditional perturbation techniques. These metrics enable effective comparisons both within and across groups of techniques.
\begin{figure}[!ht]
\centering
\includegraphics[width=0.8\textwidth]{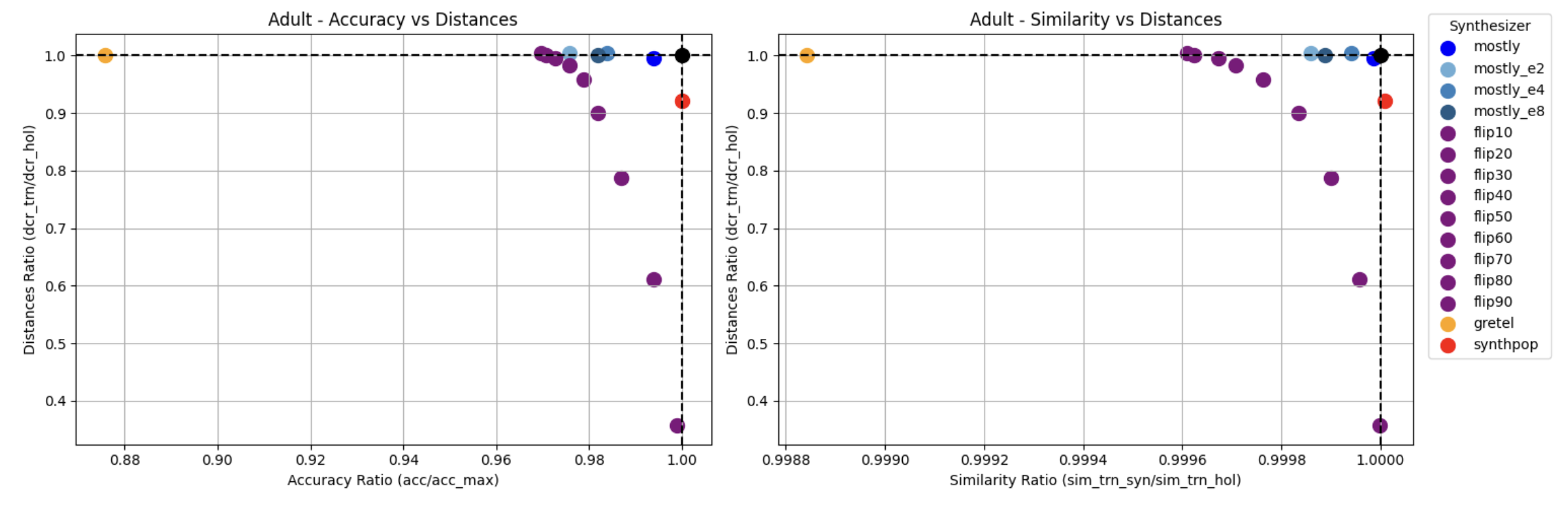}
\caption{Visualizing the fidelity-privacy tradeoff of different synthesizers using the UCI Adult Census dataset. Accuracy ratio $\text{acc}/\text{acc}_{\text{max}}$ (left) and similarity ratio $\text{sim}_{\text{trn,sim}}/\text{sim}_{\text{trn,hol}}$ (right) across different generative and perturbation techniques. \textbf{mostly}: Synthetic data generated using the Synthetic Data SDK \cite{synthetic_data_sdk}, with default model and training parameters. \textbf{mostly\_eX}: same as \textit{mostly} but training is stopped after $X$ epochs. \textbf{flipK}: Perturbed dataset where each cell is replaced with a value from a randomly selected record with probability K\%. \textbf{synthpop} \cite{nowok2016synthpop} and \textbf{gretel} \cite{gretel}: Other open-source synthesizers.}
\label{fig:similarity}
\end{figure}

\section{Conclusion}

The increasing adoption of generative models for structured data underscores the critical need for interpretable, standardized, and open-sourced tools for synthetic data quality assessment. In response, we have introduced the framework \texttt{ mostlyai-qa}, a versatile and empirically grounded Python framework that simultaneously quantifies utility and privacy protection of synthetic data. By supporting heterogeneous data structures and providing holdout-based benchmarking, the framework makes it possible to perform comparisons across synthetic data synthesizers and promotes methodological transparency. We anticipate that the framework will support both practitioners and researchers in the evaluation of synthetic data pipelines, contribute to reproducibility in generative data science, and help to standardize evaluation frameworks in this field.

\newpage
\section*{Acknowledgements}

We wish to convey our sincere appreciation to Tobias Hann, Radu Rogojanu, Lukasz Kolodziejczyk, Michael Druk, Shuang Wu, Alex Ichim, Ivona Krchova, among others, for their essential contributions to the development of the framework. Furthermore, our gratitude extends to the community of users whose ongoing feedback and support have been paramount in the refinement and advancement of this framework. Their insights have been instrumental in tailoring the framework to better address the requirements of practitioners engaged in synthetic data quality assessment. We also recognize the developers and maintainers of the open-source libraries upon which the framework relies. Specifically, we appreciate the efforts of the teams responsible for \texttt{plotly}, \texttt{scikit-learn} and \texttt{transformers}.

\newpage

\bibliographystyle{unsrt}  
\bibliography{paper}

\newpage

\appendix

\section{Summary of Evaluation Metrics}

\begin{itemize}
  \item \textbf{Accuracy}: Accuracy is defined as (100\% - Total Variation Distance), for each distribution, and then averaged across.
  \begin{itemize}
    \item \texttt{overall}: Overall accuracy of synthetic data, i.e. average across univariate, bivariate and coherence.
    \item \texttt{univariate}: Average accuracy of discretized univariate distributions.
    \item \texttt{bivariate}: Average accuracy of discretized bivariate distributions.
    \item \texttt{coherence}: Average accuracy of discretized coherence distributions. Only applicable for sequential data.
    \item \texttt{overall\_max}: Expected overall accuracy of a same-sized holdout. Serves as reference for \texttt{overall}.
    \item \texttt{univariate\_max}: Expected univariate accuracy of a same-sized holdout. Serves as reference for \texttt{univariate}.
    \item \texttt{bivariate\_max}: Expected bivariate accuracy of a same-sized holdout. Serves as reference for \texttt{bivariate}.
    \item \texttt{coherence\_max}: Expected coherence accuracy of a same-sized holdout. Serves as reference for \texttt{coherence}.
  \end{itemize}
  \item \textbf{Similarity}: All similarity metrics are calculated within an embedding space.
  \begin{itemize}
    \item \texttt{cosine\_\allowbreak similarity\_\allowbreak training\_\allowbreak synthetic}: Cosine similarity between training and synthetic centroids.
    \item \texttt{cosine\_\allowbreak similarity\_\allowbreak training\_\allowbreak holdout}: Cosine similarity between training and holdout centroids. Serves as reference for \texttt{cosine\_\allowbreak similarity\_\allowbreak training\_\allowbreak synthetic}.
    \item \texttt{discriminator\_\allowbreak auc\_\allowbreak training\_\allowbreak synthetic}: Cross-validated AUC of a discriminative model to distinguish between training and synthetic samples.
    \item \texttt{discriminator\_\allowbreak auc\_\allowbreak training\_\allowbreak holdout}: Cross-validated AUC of a discriminative model to distinguish between training and holdout samples. Serves as reference for \texttt{discriminator\_\allowbreak auc\_\allowbreak training\_\allowbreak synthetic}.
  \end{itemize}
  \item \textbf{Distances}: All distance metrics are calculated within an embedding space. An equal number of training and holdout samples is considered.
  \begin{itemize}
    \item \texttt{ims\_training}: Share of synthetic samples that are identical to a training sample.
    \item \texttt{ims\_holdout}: Share of synthetic samples that are identical to a holdout sample. Serves as reference for \texttt{ims\_training}.
    \item \texttt{dcr\_training}: Average L2 nearest-neighbor distance between synthetic and training samples.
    \item \texttt{dcr\_holdout}: Average L2 nearest-neighbor distance between synthetic and holdout samples. Serves as reference for \texttt{dcr\_training}.
    \item \texttt{dcr\_share}: The share of synthetic samples that are closer to a training sample than to a holdout sample. This shall not be significantly larger than 50\%.
  \end{itemize}
\end{itemize}

\section{Framework Installation and Example Usage}

The presented framework for evaluating the quality of synthetic data requires Python version 3.10 or later, and can be easily installed using \texttt{pip}:
\begin{lstlisting}
pip install -U mostlyai-qa
\end{lstlisting}
Once installed, its main interface is the `report`, which expects the data samples to be provided as \texttt{pandas} DataFrames:
\begin{lstlisting}
from mostlyai import qa

# analyze single-table data
report_path, metrics = qa.report(
    syn_tgt_data=synthetic_df,
    trn_tgt_data=training_df,
    hol_tgt_data=holdout_df,
)

# analyze sequential data with context
report_path, metrics = qa.report(
    syn_tgt_data=synthetic_df,
    trn_tgt_data=training_df,
    hol_tgt_data=holdout_df,
    syn_ctx_data=synthetic_context_df,
    trn_ctx_data=training_context_df,
    hol_ctx_data=holdout_context_df,
    ctx_primary_key="id",
    tgt_context_key="user_id",
)
\end{lstlisting}
Additional usage examples, along with their corresponding HTML reports, are available in the GitHub repository \hyperlink{https://github.com/mostly-ai/mostlyai-qa}{https://github.com/mostly-ai/mostlyai-qa}.

\end{document}